\documentclass[11pt,a4paper]{article}
\usepackage[hyperref]{acl2019}
\usepackage{times}
\usepackage{latexsym}

\usepackage{url}

\usepackage{graphicx}
\usepackage{multirow}
\usepackage{multicol}

\aclfinalcopy 

\title{UdS Submission for the WMT 19 Automatic Post-Editing Task}

\author{Hongfei Xu \\
  Saarland University \\
  DFKI \\
  \texttt{hfxunlp@foxmail.com} \\\And
  Qiuhui Liu \\
  China Mobile Online Services \\
  \texttt{liuqiuhui@cmos.chinamobile.com} \\\And
  Josef van Genabith \\
  Saarland University \\
  DFKI \\
  \texttt{josef.van\_genabith@dfki.de} \\
  }

\date{}

\begin{document}
\maketitle
\begin{abstract}
  In this paper, we describe our submission to the English-German APE shared task at WMT 2019. We utilize and adapt an NMT architecture originally developed for exploiting context information to APE, implement this in our own transformer model and explore joint training of the APE task with a de-noising encoder.
\end{abstract}

\section{Introduction}

The Automatic Post-Editing (APE) task is to automatically correct errors in machine translation outputs. This paper describes our submission to the English-German APE shared task at WMT 2019. Based on recent research on the APE task \citep{junczysdowmunt2018ape} and an architecture for the utilization of document-level context information in neural machine translation \citep{zhang2018improving}, we re-implement a multi-source transformer model for the task. Inspired by \citet{cheng2018towards}, we try to train a more robust model by introducing a multi-task learning approach which jointly trains APE with a de-noising encoder.

We made use of the artificial eScape data set \citep{Matteo2018ESCAPE} provided for the task, since the multi-source transformer model contains a large number of parameters and training with large amounts of supplementary synthetic data can help regularize its parameters and make the model more general. We then tested the BLEU scores between machine translation results and corresponding gold standard post-editing results on the original development set, the training set and the synthetic data as shown in Table \ref{tab:datableu}.

\begin{table}[htbp]
  \centering
    \begin{tabular}{|r|r|r|}
    \hline
    \multicolumn{1}{|l|}{dev} & \multicolumn{1}{l|}{train} & \multicolumn{1}{l|}{eScape} \\
    \hline
    77.15  & 77.42  & 37.68  \\
    \hline
    \end{tabular}%
  \caption{BLEU Scores of Data Sets}
  \label{tab:datableu}%
\end{table}%

Table \ref{tab:datableu} shows that there is a significant gap between the synthetic eScape data set \citep{Matteo2018ESCAPE} and the real-life data sets (the development set and the original training set from post-editors), potentially because \citet{Matteo2018ESCAPE} generated the data set in a different way compared to \citet{junczysdowmunt2016log} and very few post-editing actions are normally required due to the good translation quality of neural machine translation \citep{bahdanau2014neural, gehring2017convolutional, vaswani2017attention} which significantly reduces errors in machine translation results and makes the post-editing results quite similar to raw machine translation outputs.

\section{Our Approach}

\begin{figure*}[htbp]
  \centering
  \includegraphics[scale=0.50]{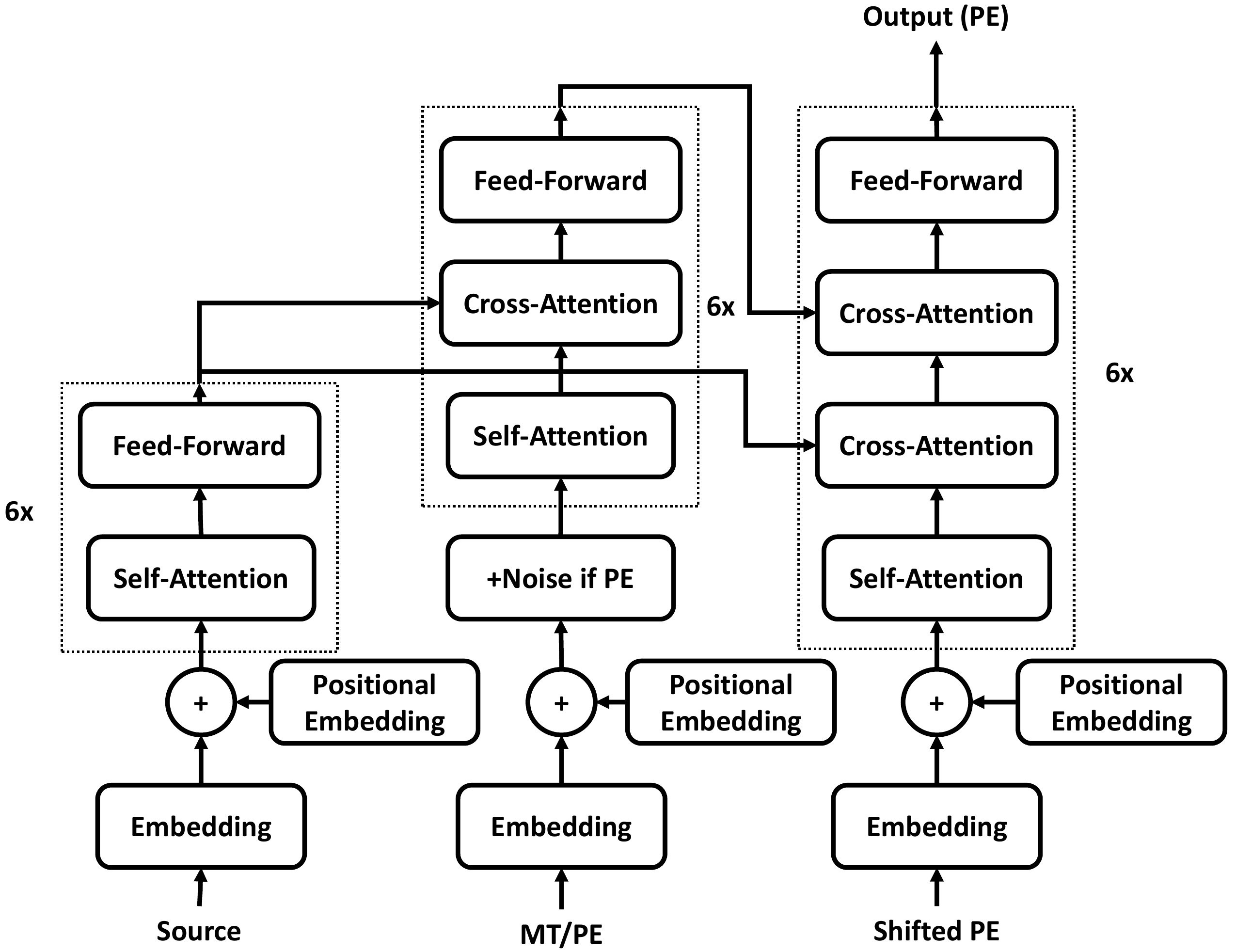}\\
  \caption{Our Transformer-Based Multi-Source Model for the APE Task}\label{fig:model}
\end{figure*}

We simplify and employ a multi-source transformer model \citep{zhang2018improving} for the APE task, and try to train a more robust model through multi-task learning.

\subsection{Our Model}

The transformer-based model proposed by \citet{zhang2018improving} for utilizing document-level context information in neural machine translation has two source inputs which can also be a source sentence along with the corresponding machine translation output and therefore caters for the requirements of APE. Since both source sentence and machine translation outputs are important for the APE task \citep{pal2016neural, vuhaffari2018automatic}, we remove the context gate used to restrict the information flow from the first input to the final output in their architecture, and obtain the model we used for our submission shown in Figure \ref{fig:model}.

The model first encodes the given source sentence with stacked self-attention layers, then ``post-edits'' the corresponding machine translation result through repetitively encoding the machine translation result (with a self-attention layer), attending to the source sentence (with a cross-attention layer) and processing the collected information (with a feed-forward neural network). Finally, the decoder attends to representations of the source sentence and the machine translation result and generates the post-editing result.

Compared to the multi-source transformer model used by \citet{junczysdowmunt2018ape}, this architecture has one more cross-attention module in the encoder for machine translation outputs to attend to the source input which makes the parameter sharing of layers between two encoders impossible, but we think this cross-attention module can help the de-noising task. The embedding of source, machine translation outputs and post-editing results is still shared as \citet{junczysdowmunt2018ape} advised.

\subsection{Joint Training with De-noising Encoder}

Table \ref{tab:datableu} shows a considerable difference between the synthetic data set \citep{Matteo2018ESCAPE} and the real data set. To enable the model to handle more kinds of errors, we simulate new ``machine translation outputs'' through adding noise to the corresponding post-editing results. Following \citet{cheng2018towards}, we add noise directly to the look-up embedding of post-editing results instead of manipulating post-editing sequences.

Since the transformer \citep{vaswani2017attention} does not apply any weight regularization, we assume that the model can easily learn to reduce noise by enlarging weights, and propose to add adaptive noise to the embedding:
\begin{equation}
    em{b_{out}} = emb + strength * \overline {abs(emb)} * N
\end{equation}

where $emb$ is the embedding matrix, $strength$ is a number between $[0.0, +\infty)$ to control the strength of noise, $N$ is the noise matrix of the same shape as $emb$. We explore both standard Gaussian distribution and uniform distribution of $[-1.0, -1.0]$ as $N$. In this way the noise will automatically grow with the growing embedding weights.

Given that the transformer translation model \citep{vaswani2017attention} incorporates word order information through adding positional embedding to word embedding, we add noise to the combined embedding. In this case, the noise can both affect the word embedding (replacing words with their synonyms) and positional embedding (swapping word orders).

During training, we use the same model, and achieve joint training by randomly varying inputs: the inputs for the APE task are \{source, mt, pe\}, while those for the de-noising encoder task are \{source, pe+noise, pe\} where ``source'', ``mt'' and ``pe'' stand for the source sentence, the corresponding output from the machine translation system and the correct post-editing result. The final loss for joint training is:
\begin{equation}
    loss = \lambda *los{s_{{\rm{ape}}}} + (1 - \lambda )*los{s_{{\rm{de - noising}}}}
\end{equation}

i.e. the loss between the APE task and the de-noising encoder task are balanced by $\lambda$ in this way.

\section{Experiments}

We implemented our approaches based on the Neutron implementation \citep{xu2019neutron} for transformer-based neural machine translation.

\subsection{Data and Settings}

We only participated in the English to German task, and we used both the training set provided by WMT and the synthetic eSCAPE corpus \citep{Matteo2018ESCAPE}. We first re-tokenized\footnote{using arguments: -a -no-escape} and truecased both data sets with tools provided by Moses \citep{koehn2007moses}, then cleaned the data sets with scripts ported from the Neutron implementation, and the original training set was up-sampled $20$ times as in \citep{junczysdowmunt2018ape}. We applied joint Byte-Pair Encoding \citep{sennrich2015neural} with $40k$ merge operations and $50$ as the vocabulary threshold for the BPE. We only kept sentences with a max of $256$ sub-word tokens for training, and obtained a training set of about $6.5M$ triples with a shared vocabulary of $42476$. We did not apply any domain adaptation approach for our submission considering that \citep{junczysdowmunt2018ape} shows few improvements, but advanced domain adaption \citep{wang2017sentence} or fine-tuning \citep{Luong15iwslt} methods may still bring some improvements. The training set was shuffled for each training epoch.

Like \citet{junczysdowmunt2018ape}, all embedding matrices were bound with the weight of the classifier. But for tokens which in fact do never appear in post-editing outputs in the shared vocabulary, we additionally remove their weights in the label smoothing loss and set corresponding biases in the decoder classifier to $-10^{32}$.

Unlike \citet{zhang2018improving}, the source encoder, the machine translation encoder and the decoder had $6$ layers. The hidden dimension of the position-wise feed-forward neural network was $2048$, the embedding dimension and the multi-head attention dimension were $512$. We used a dropout probability of $0.1$, and employed label smoothing \citep{Szegedy2016Rethinking} value of 0.1. We used the Adam optimizer \citep{Kingma15Adam} with $0.9$, $0.98$ and $10^{-9}$ as $\beta_{1}$, $\beta_{2}$ and $\epsilon$. The learning rate schedule from \citet{vaswani2017attention} with $8,000$ as the number of warm-up steps\footnote{\url{https://github.com/tensorflow/tensor2tensor/blob/master/tensor2tensor/models/transformer.py\#L1623}.} was applied. We trained our models for only $8$ epochs with at least $25k$ post-editing tokens in a batch, since we observed over-fitting afterwards. For the other hyper parameters, we used the same as the transformer base model \citep{vaswani2017attention}.

During training, we kept the last $20$ checkpoints saved with an interval of $1,500$ training steps \citep{vaswani2017attention, zhang2018accelerating}, and obtained $4$ models for each run through averaging every $5$ adjacent checkpoints.

For joint training, we simply used $0.2$ as the strength of noise ($strength$), and $0.5$ as $\lambda$ for joint training. Other values may provide better performance, but we did not have sufficient time to try this for our submission.

During decoding, we used a beam size of $4$ without any length penalty.

\subsection{Results}

We first evaluated case-sensitive BLEU scores\footnote{\url{https://github.com/moses-smt/mosesdecoder/blob/master/scripts/generic/multi-bleu.perl}.} on the development set, and results of all our approaches and baselines are shown in Table \ref{tab:bleudev}.

``MT as PE'' is the do-nothing baseline which takes the machine translation outputs directly as post-editing results. ``Processed MT'' is the machine translation outputs through pre-processing (re-tokenizing and truecasing) and post-processing (de-truecasing and re-tokenizing without ``-a'' argument\footnote{``-a'' indicates tokenizing in the aggressive mode, which normally helps reduce vocabulary size. The official data sets were tokenized without this argument, so we have to recover our post-editing outputs.}) but without APE. ``Base'', ``Gaussian'' and ``Uniform'' stand for our model trained only for the APE task, jointly trained with Gaussian noise and uniform noise, respectively. We reported the minimum and the maximum BLEU scores of the 4 averaged models for each experiment. ``Ensemble x5'' is the ensemble of 5 models from joint training, 4 of which were averaged models with highest BLEU scores on the development set, another one was the model saved for each training epoch with lowest validation perplexity.

\begin{table}[htbp]
  \centering
    \begin{tabular}{|l|l|}
    \hline
    Models & BLEU \\
    \hline
    MT as PE & \multicolumn{1}{r|}{76.76} \\
    \hline
    Processed MT & \multicolumn{1}{r|}{76.61} \\
    \hline
    Base  & 76.91 $\sim$ 77.13 \\
    \hline
    Gaussian & 76.94 $\sim$ 77.08 \\
    \hline
    Uniform & 77.01 $\sim$ 77.10 \\
    \hline
    Ensemble x5 & \multicolumn{1}{r|}{\textbf{77.22}} \\
    \hline
    \end{tabular}%
  \caption{BLEU Scores on the Development Set}
  \label{tab:bleudev}%
\end{table}%

Table \ref{tab:bleudev} shows that the performance got slightly hurt (comparing ``Processed MT'' with ``MT as PE'') with pre-processing and post-processing procedures which are normally applied in training seq2seq models for reducing vocabulary size. The multi-source transformer (Base) model achieved the highest single model BLEU score without joint training with the de-noising encoder task. We think this is perhaps because there is a gap between the generated machine translation outputs with noise and the real world machine translation outputs, which biased the training.

Even with the ensembled model, our APE approach does not significantly improve machine translation outputs measured in BLEU (+0.46). We think human post-editing results may contain valuable information to guide neural machine translation models in some way like Reinforcement-Learning, but unfortunately, due to the high quality of the original neural machine translation output, only a small part of the real training data in the APE task are actually corrections from post editors, and most data are generated from the neural machine translation system, which makes it like adversarial training of neural machine translation \citep{yang2018improving} or multi-pass decoding \citep{geng2018adaptive}.

All our submissions were made by jointly trained models because the performance gap between the best and the worst model of jointly trained models is smaller, which means that jointly trained models may have smaller variance.

Results on the test set from the APE shared task organizers are shown in Table \ref{tab:test}. Even the ensemble of 5 models did not result in significant differences especially in BLEU scores.

\begin{table}[htbp]
  \centering
    \begin{tabular}{|l|r|r|}
    \hline
    Models & \multicolumn{1}{l|}{TER} & \multicolumn{1}{l|}{BLEU} \\
    \hline
    MT as PE & 16.84  & 74.73  \\
    \hline
    Gaussian & 16.79  & 75.03  \\
    \hline
    Uniform & 16.80  & 75.03  \\
    \hline
    Ensemble x5 & 16.77  & 75.03  \\
    \hline
    \end{tabular}%
  \caption{Results on the Test Set}
  \label{tab:test}%
\end{table}%

\section{Related Work}

\citet{pal2016neural} applied a multi-source sequence-to-sequence neural model for APE, and \citet{vuhaffari2018automatic} jointly trained machine translation with the post editing sequence prediction task \citep{berard2017WMT}. Though all previous approaches get significant improvements over Statistical Machine Translation outputs, benefits with APE on top of Neural Machine Translation outputs are not very significant \citep{chatterjee2018WMT}.

On the other hand, advanced neural machine translation approaches may also improve the APE task, such as: combining advances of the recurrent decoder \citep{chen2018best}, the Evolved Transformer architecture \citep{david2019the}, Layer Aggregation \citep{dou2018exploiting} and Dynamic Convolution structures \citep{wu2018pay}.

\section{Conclusion}

In this paper, we described details of our approaches for our submission to the WMT 19 APE task. We borrowed a multi-source transformer model from the context-dependent machine translation task and applied joint training with a de-noising encoder task for our submission.

\section*{Acknowledgments}
Hongfei Xu is supported by a doctoral grant from China Scholarship Council ([2018]3101, 201807040056). This work is supported by the German Federal Ministry of Education and Research (BMBF) under the funding code 01IW17001 (Deeplee). We thank the anonymous reviewers for their instructive comments.

\bibliography{acl2019}
\bibliographystyle{acl_natbib}

\end{document}